%% file: root.tex
\documentclass[letterpaper, 10 pt, conference]{ieeeconf}

\usepackage{times}
\usepackage{graphicx}
\usepackage{wrapfig}
\usepackage{xfrac}

\usepackage{multicol}
\usepackage{amsmath}
\usepackage{siunitx}
\usepackage{circuitikz}
\usetikzlibrary{patterns,hobby,decorations.pathmorphing}

\usepackage{multirow}

\usepackage{makecell}

\usepackage{booktabs}

\usepackage[hidelinks]{hyperref}
\usepackage{cleveref}
\Crefformat{figure}{#2Fig.~#1#3}
\Crefmultiformat{figure}{Figs.~#2#1#3}{ and~#2#1#3}{, #2#1#3}{ and~#2#1#3}
\crefformat{footnote}{#2\footnotemark[#1]#3}

\IEEEoverridecommandlockouts

\title{\LARGE \bf
Sim2Real for Soft Robotic Fish via Differentiable Simulation
}

\author{John Z. Zhang$^{1,2}$, Yu Zhang$^{1}$, Pingchuan Ma$^{3}$,  Elvis Nava$^{1,4}$, Tao Du$^{3}$, Philip Arm$^{1}$, \\ Wojciech Matusik$^{3}$, Robert K. Katzschmann$^{1,4}$
\thanks{*We are grateful for funding received by the ETH AI Center, and the Defense Advanced Research Projects Agency.}%
\thanks{$^{1}$Soft Robotics Lab, ETH Zurich, Switzerland}%
\thanks{$^{2}$Mechanical Engineering, MIT, USA}%
\thanks{$^{3}$Computer Science and AI Lab, MIT, USA}%
\thanks{$^{4}$ETH AI Center, ETH Zurich, Switzerland}
\thanks{Corresponding author: Robert K. Katzschmann, {\tt\footnotesize 
\href{mailto:rkk@ethz.ch}{rkk@ethz.ch}}}
}

\begin{document}

\maketitle
\thispagestyle{empty}
\pagestyle{empty}

\begin{abstract}
Accurate simulation of soft mechanisms under dynamic actuation is critical for the design of soft robots.
We address this gap with our differentiable simulation tool by learning the material parameters of our soft robotic fish.
On the example of a soft robotic fish, we demonstrate an experimentally-verified, fast optimization pipeline for learning the material parameters from quasi-static data via differentiable simulation and apply it to the prediction of dynamic performance.
Our method identifies physically plausible Young’s moduli for various soft silicone elastomers and stiff acetal copolymers used in creation of our three different robotic fish tail designs. We show that our method is compatible with varying internal geometry of the actuators, such as the number of hollow cavities.
Our framework allows high fidelity prediction of dynamic behavior for composite bi-morph bending structures in real hardware to millimeter-accuracy and within $3\%$ error normalized to actuator length.
We provide a differentiable and robust estimate of the thrust force using a neural network thrust predictor; this estimate allows for accurate modeling of our experimental setup measuring bollard pull. 
This work presents a prototypical hardware and simulation problem solved using our differentiable framework; the framework can be applied to higher dimensional parameter inference, learning control policies, and computational design due to its differentiable character.
\end{abstract}

\input{main/intro}
\input{main/related}
\input{main/system}
\input{main/modeling}
\input{main/experiments}
\input{main/identification}
\input{main/results}
\input{main/conclusion}
\input{main/appendix}


\bibliographystyle{IEEEtran}
\bibliography{references}

\end{document}

%% file: main/intro.tex
\section{Introduction}

Soft robots are favored over traditional hard robots in a growing list of scenarios, including bio-inspired design, conformal gripping, co-bot situations, and more. While the design and simulation of hard robots are well understood and extensively shown thanks to decades of development, the same simulation task for compliant structures and their interaction with the environment is an ongoing challenge. 

Although many simulation schemes exist, none have demonstrated sufficient sim2real matching through verifiable experiments for soft robots in a hydrodynamic situation such as a robotic fish. The high dimensionality of soft mechanisms and the complex physics of fluid-structure interactions (FSI) make this problem particularly difficult. One critical challenge in producing high accuracy simulations is the precise measurement and manual tuning needed to dial-in the material parameters of soft mechanisms. The problem has recently become more tractable because of advances in physics-based simulation, such as differentiability and data-driven approaches. 

\begin{figure}[tb]
    \centering
    \includegraphics[width=\linewidth]{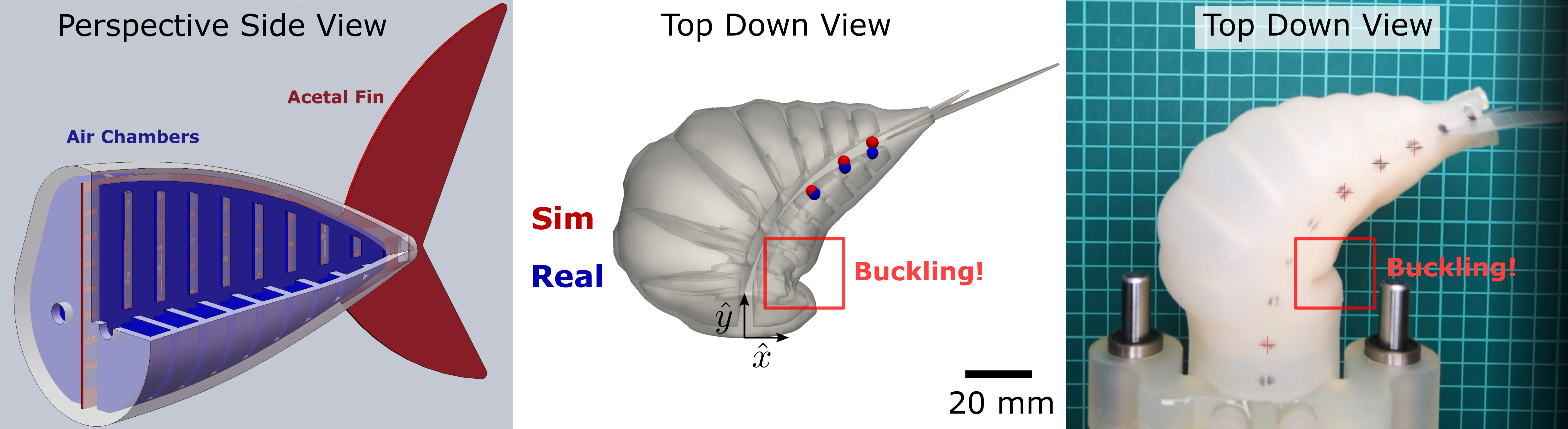}
    \caption{\textit{Left:} A model of a pneumatic fish tail (\emph{Nemo}). \textit{Middle:} The same fish tail inflated to \SI{250}{mbar} in simulation. \textit{Right:} The actual hardware of the fish tail in an experimental setup for measuring the deformation and the thrust of the fish in air and underwater. Our framework is capable of reproducing the buckling effect in simulation, a valuable result.}
    \label{fig:pneumatic_fish_tail}
    \vspace{-8pt}
\end{figure}

We present a differentiable simulation framework that can be used to accurately predict the deformation of a soft pneumatic actuator and identify its material properties through gradient-based optimization. Three designs of a pneumatically-actuated fish tail are used as benchmark (see \Cref{tab:fish}) since composite bi-morph structures are difficult to simulate due to the disparate Young's moduli and aspect ratios of the deformable body and stiffer spine. The deformable body of the fish tail is cast out of silicone elastomer and has a flexible spine made of an acetal plastic sheet in the center (see \Cref{fig:pneumatic_fish_tail}). .

Our framework is capable of learning physically-plausible material parameters, i.e., Young's modulus and Poisson's ratio, using only a quasistatic data set without extensive material testing. We demonstrate that after optimization we can produce accurate simulation results compared to dynamic data collected in our hardware setup to within millimeter accuracy or within $3\%$ max error normalized to the actuator length. The fast system identification is achieved using our finite element method (FEM)~\cite{du2021diffpd,ma2021diffaqua}, which combines projective dynamics~\cite{bouaziz2014projective} and differentiable simulation~\cite{hu2019difftaichi}.

To verify our simulation results, we developed an experimental setup for collecting position, pressure, and force data using marker tracking, a pneumatic valve array, and a load cell. The raw data is synchronized and analyzed in MATLAB.

Furthermore, we are able to reproduce the measured thrust in our bollard-pull type experiment using a simple neural network predictor that can be integrated in our differentiable simulator. Our model is capable of predicting thrust performance in actuation signals not seen in the training set and preserves differentiability of the simulator. Our aim is to provide straightforward sim2real methods for roboticists that are within reach: such methods could be used confidently for design optimization if they are verifiable by real data.

In this paper, we contribute:
\begin{itemize}
    \item a simulation framework for a soft pneumatic bending actuator consisting of disparate materials and geometries;
    \item a system identification method that uses differentiable simulation and gradient-based optimization to accurately learn material parameters of two isotropic corotated materials;
    \item a data-driven hydrodynamics model using a neural network as a simple predictor of the thrust force generated by the fish; and
    \item experimental verification of our simulation results using a hardware setup suitable for data acquisition in both air and water.
\end{itemize}

%% file: main/related.tex
\section{Related Work}
\label{related_work}

\subsection{Soft underwater robots}
Soft robots are difficult to optimally design and control when compared to their rigid counterparts due to the infinite dimensionality of their compliant structures.
Due to this modeling complexity, an experienced designer must hand craft each design guided by intuition, experiments, and approximate models.
Marchese et al. offer approaches to designing and fabricating soft fluidic elastomer robots, the type of robot we are also using in this work~\cite{marchese2015recipe}.
Katzschmann et al. present the design, fabrication, control, and testing of a soft robotic fish with interior cavities that is hydraulically actuated. Their manually designed robot can swim at multiple depths and record aquatic life in the ocean~\cite{katzschmann2016hydraulic,katzschmann2018exploration}. 
Zhu et al. manually optimize the swimming performance of their robotic fish, Tunabot~\cite{zhu2019tuna}. The authors measured kinematics, speed, and power at increasing flapping frequencies to quantify swimming performance and find agreement in performance between real fish and their Tunabot over a wide range of frequencies.
Zheng et al. propose to design soft robots by pre-checking controllability during the numerical design phase~\cite{zheng2019controllability}. FEM is used to model the dynamics of cable-driven parallel soft robot and a differential geometric method is applied to analyze the controllability of the points of interest.
Katzschmann et al.~\cite{katzschmann2019dynamically} manually tweak the material parameters of their reduced-order FEM~\cite{thieffry2018control} with an experimental soft robotic arm to perform dynamic closed-loop control.
Van et al. present a DC motor driven soft robotic fish which is optimized for speed and efficiency based on experimental, numerical and theoretical investigation into oscillating propulsion~\cite{van2020biomimetic}.
Wolf et al. use a pneumatically-actuated fish-like stationary model to investigate how parameters like stiffness, strength, and frequency affect thrust force generation~\cite{wolf2020fish}. Wolf et al. measure thrust, side forces, and torques generated during propulsion and use a statistical linear model to examine the effects of parameter combinations on thrust generation; they show that both stiffness and frequency substantially affect swimming kinematics.
We are not aware of any work that uses a fast differentiable FEM simulation environment to learn material parameters for soft robotic fish using a bollard-pull style experimental setup.

\subsection{Differentiable soft-body simulators}
Our work is also relevant to the recent developments of robotic simulators, particularly for soft robots.
Geilinger et al. \cite{geilinger2020add} present a differentiable multi-body dynamics solver that is able to handle frictional contact for rigid and deformable objects.
Coevoet et al. \cite{coevoet2017software} notably present a non-differentiable framework for modeling, simulation, and control of soft-bodied robots using continuum mechanics for modeling the robotic components and using Lagrange multipliers for boundary conditions like actuators and contacts.
Most related to our work are the recent works on differentiable soft-body and fluid simulators~\cite{du2020stokes,du2021diffpd,hahn2019real2sim,hu2019difftaichi,hu2019chainqueen,huang2021plasticine,ma2021diffaqua}. 
These papers develop numerical methods for computing gradients in a traditional simulators. Furthermore, they demonstrate the power of gradient information in robotics applications, e.g., system identification or trajectory optimization. Most of the works present simulation results only, with ChainQueen~\cite{hu2019chainqueen} and Real2Sim~\cite{hahn2019real2sim} being two notable exceptions that discuss real-world soft-robot applications.
Notably, \cite{hahn2019real2sim} optimizes visco-elastic material parameters of a finite element simulation to approximate the dynamic deformations of real-world soft objects, such as an open-loop controlled tendon-driven crawling robot.
Bern et al. \cite{bern2020soft} have also demonstrated the use of differentiable simulation to learn from a quasi-static data set for the purpose of optimizing open-loop control inputs.
Dubied et al. \cite{dubied2022sim} is the most recent example that demonstrates sim2real matching for a soft robotic fish tail, shows system identification on a passive structure for just the Young's modulus, and investigates the mismatch in damping between reality and simulation. In this previous work, the fish tail actuation is simulated using a simplified muscle model and only one design is shown whereas in this paper, the pressure boundary condition is simulated exactly as fabricated for each pneumatic chamber geometry for three different designs. Simulating the pneumatic chambers improves accuracy and allows for physically-plausible Young's moduli and Poisson ratios to be identified. In this current work, we further demonstrate that the gradient-based optimization can be carried out to higher dimensional design spaces that include more than one material parameter.

\subsection{Hydrodynamic Surrogates}
For underwater soft robots, the challenge of simulation is exacerbated by the hydrodynamic interaction with the soft body.
Several previous works tackle the fluid-structure interaction problem through different methods, including heuristic hydrodynamics~\cite{du2021underwater,ma2021diffaqua,min2019softcon}, physically-informed neural network approaches~\cite{wandel_learning_2021}, and data-driven learning approaches~\cite{chen2018neural}.

Compared to these previous methods that simulate underwater soft robots such as~\cite{du2021underwater}, our work models pneumatic actuation using the exact chamber geometry rather than artificial muscles facilitating greater accuracy at large deformations (see \Cref{fig:pneumatic_fish_tail}), uses a neural network thrust predictor rather than approximate analytical or heuristic hydrodynamics, and presents a more sophisticated hardware pipeline that can be used to validate simulation. 

%% file: main/system.tex
\section{System Overview}
\label{system_overview}

\begin{figure}[tb]
    \centering
    \includegraphics[width=0.48\textwidth]{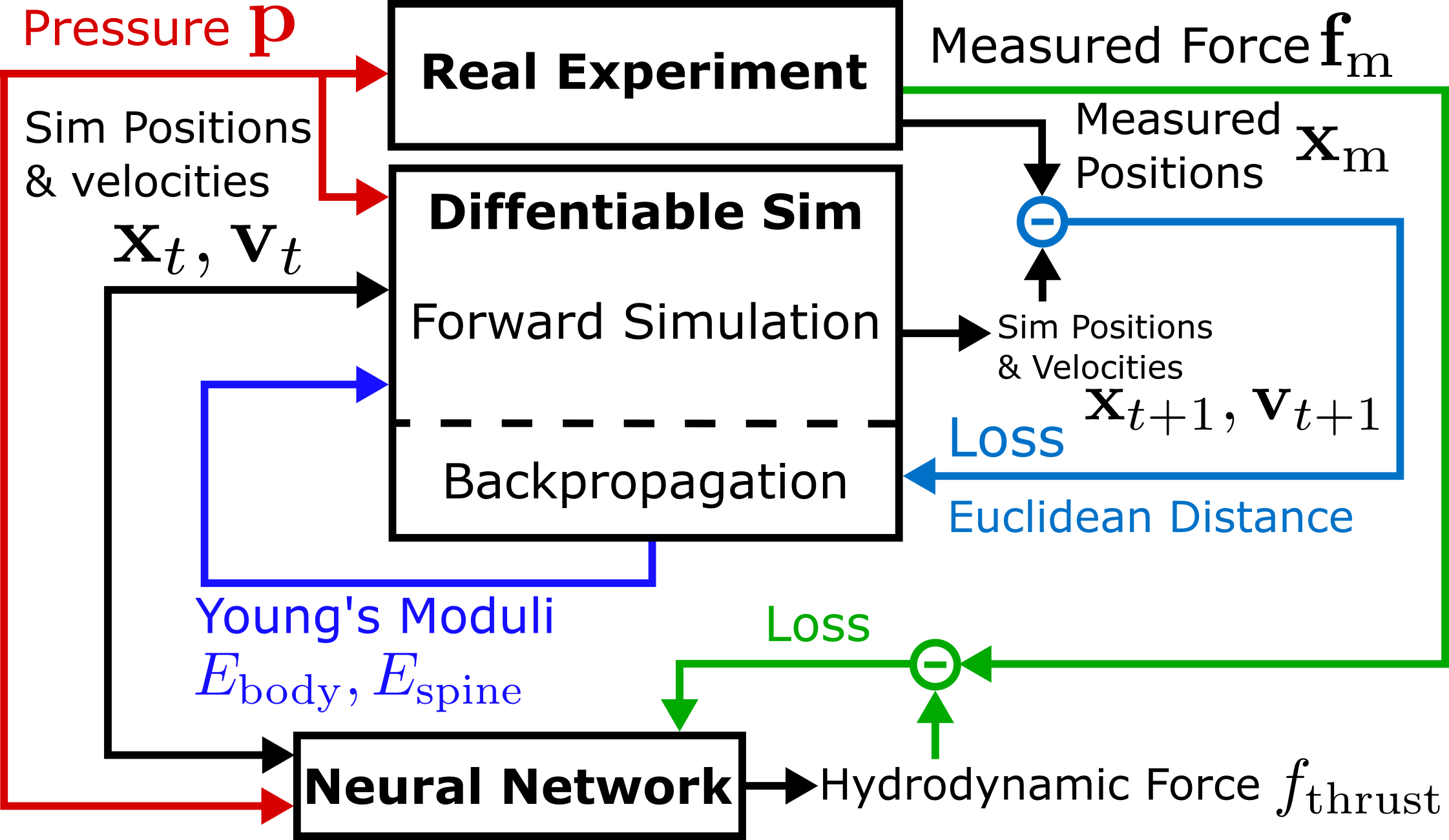}
    \caption{Flow diagram of our differentiable simulation and learning pipeline. To learn the material parameters, we compute the loss as the Euclidean distance between the measured marker positions $\mathbf{x}_\textrm{m}$ and the simulated position data $\mathbf{x}_t$ and then minimize this loss using gradient-search enabled by our differentiable FEM simulation. The marker positions are 2D in-plane measurements primarily due to lateral displacement during flapping. Our neural network thrust predictor takes as input the positions and pressure at a previous time step and outputs the thrust force for the next time step.}
    \label{fig:overview}
\end{figure}

Our pipeline (\Cref{fig:overview}) learns the material parameters of a soft robotic fish (see \Cref{fig:pneumatic_fish_tail}) using differentiable simulation and measurements from quasistatic experiments. Thrust force is simulated using a data-driven neural network predictor. We demonstrate that sim2real matching is good within $3\%$ positional error normalized to actuator length for our soft robotic fish prototypes even under significant bending.

In \Cref{simulation}, forward simulation is accomplished using implicit Euler time stepping and FEM spatial discretization. We leverage projective dynamics for a substantial speed up in computation\,\cite{du2021diffpd}.
We describe our bollard-pull style experimental setup in \Cref{experimental_setup}.
In \Cref{system identification}, the material properties of the body and spine are specified by two different corotated materials stitched together at the boundary.
We verify the results (\Cref{fig:system_id}) of our system identification by comparing simulations to measurements from a gamut of dynamic experiments at varying amplitudes and frequencies. Finally, in \Cref{hydrodynamics} we describe the architecture and training of our thrust predictor network.

%% file: main/modeling.tex
\section{Physical Modeling}
\label{physical_modeling}

\subsection{Solid mechanics}
\label{solid_mechanics}
The deformation of a soft material can be modeled by the equations of elasticity. For the constitutive model we choose an isotropic corotated material~\cite{sifakis2012fem} since it is straightforward to implement for projective dynamics and can predict bending to within sufficient accuracy for engineering purposes.

A corotated material is defined by its energy density function using the Frobenius Norm and the trace operator,
\begin{equation}
    \Psi = \mu \left\lVert\mathbf{U} - \mathbf{I}\right\rVert^2_F + \frac{\lambda}{2}\text{tr}^2(\mathbf{U}-\mathbf{I}),
\end{equation}
where $\mathbf{U}$ is the stretch tensor, $\mathbf{I}$ is the identity, and $\mu$ and $\lambda$ are the Lam\'e parameters, which have the following relation to the Young's modulus $E$ and Poisson's ratio $\nu$ for an isotropic material,
\begin{equation}
    \mu = \frac{E}{2(1+\nu)}, \quad \lambda = \frac{E\nu}{(1+\nu)(1-2\nu)}.
\end{equation} This material energy model is popular for physical simulation and animation, but it is not the most accurate choice for closing the sim2real gap for large deformations~\cite{sifakis2012fem, adeeb2011introduction}. 

To simulate a pneumatically actuated fish tail, we use a spatially-constant time-varying pressure boundary condition on the interior surface of each chamber of the tail.

\subsection{Hydrodynamics}
\label{hydrodynamics}
The hydrodynamic effects that produce thrust and drag on a swimmer are important, however, the physics governing the fluid can be complicated due to turbulent effects and two-way coupling between the solid and fluid domains. One may choose to model water using the Navier-Stokes equations, however, in practice simulating the full FSI problem is under time constraints computationally unfeasible. For this reason, many workers in the field of computer graphics and simulation choose to use simplified heuristic hydrodynamic models such as the one presented by Min et al.~\cite{min2019softcon}. Analytical models for fish propulsion have also been studied extensively in the literature~\cite{lighthill1971large, wu1961swimming, triantafyllou2000hydrodynamics}, but these approximations require severe simplifying assumptions necessary for analytical tractability. Using the elongated-body theory (EBT) described in \Cref{sec:large_amplitude_ebt}~\cite{lighthill1971large}, the average thrust produced by a fish is approximated by
$f_\textrm{EBT} = m \hat{g}(\dot{x},\dot{y},y)$,
where $m$ is the virtual mass and $\hat{g}$ is a function of the velocity $\dot{x},\dot{y}$ and deformation $y$ of the fish tail measured in millimeters. We use EBT for comparison to our learned hydrodynamic model.

\subsection{Simulation}
\label{simulation}

\subsubsection{Discretization} 
We use our finite element method with tetrahedra to discretize the continuous spatial domain occupied by the soft fish tail. Furthermore, we implement the implicit Euler time-stepping scheme for time discretization. Formally, let $\mathbf{x}_i$ and $\mathbf{v}_i$ be vectors of size $3N$ stacking up the nodal positions and velocities from the tetrahedra at the $i$-th time step. We then rewrite the governing equations in the following discretized form
\begin{align}\label{eqn:discretization}
    \mathbf{x}_{i+1}=&\mathbf{x}_{i}+h\mathbf{v}_{i+1},\\
    \mathbf{v}_{i+1}=&\mathbf{v}_i+h\mathbf{M}^{-1}[\mathbf{f}_{ela}(\mathbf{x}_{i+1})+\mathbf{f}_{act}].
\end{align}
Here, $h$ is the time step, $M$ is the mass matrix, and $\mathbf{f}_{ela}$ and $\mathbf{f}_{act}$ represent the elastic force and the actuator force (computed by the pressure from the pneumatic actuator and the surface area of the chambers), respectively. The time integration is implicit because we make the elastic force depend on the nodal position $\mathbf{x}_{i+1}$ at the beginning of the next time step instead of the current one.

\subsubsection{Solver} 
Solving \Cref{eqn:discretization} typically requires expensive numerical computation due to its implicit time-stepping scheme. However, mature numerical tools exist since this problem has been extensively studied in computer graphics and physics simulation. In this work, we choose to use DiffPD~\cite{du2021diffpd}, a recent differentiable simulator with projective dynamics that is suitable for simulating terrestrial and underwater soft robots. We choose differentiable projective dynamics over other numerical solvers, for example traditional Newton's methods, due to its simultaneous accommodation of speed, robustness, and differentiability.

%% file: main/experiments.tex
\section{Experimental Setup}
\label{experimental_setup}

\begin{figure}
    \centering
    \includegraphics[width=\linewidth]{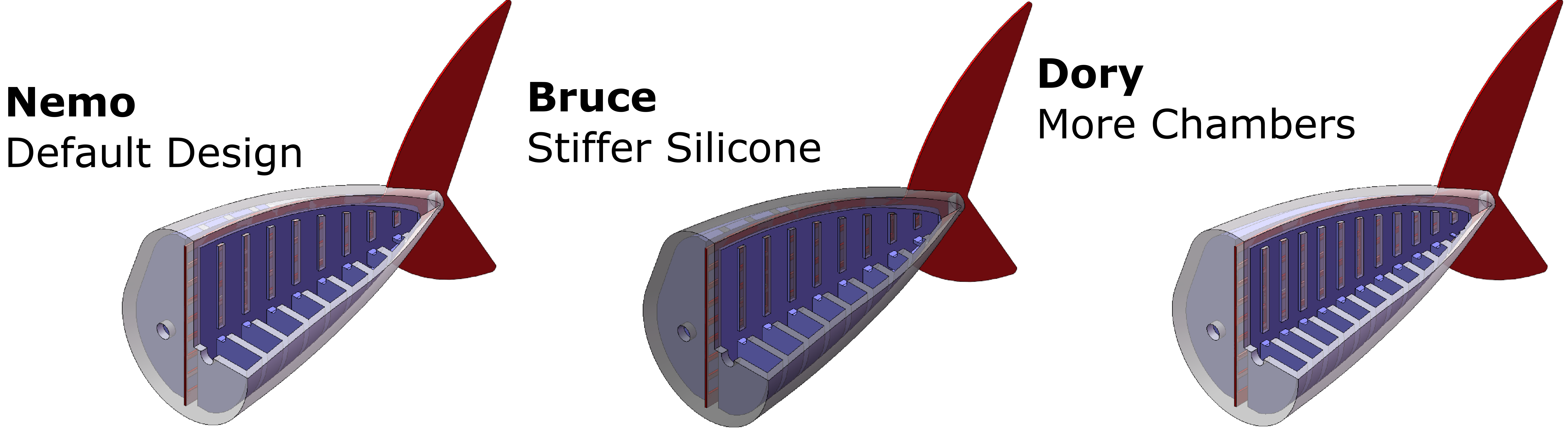}
    \caption{Three different actuator designs. \emph{Nemo} and \emph{Bruce} share the same geometry but \emph{Bruce} has a stiffer DS20 material for its body. \emph{Nemo} and \emph{Dory} share the same materials but \emph{Dory} has a greater number of air chambers.}
    \label{fig:fish_CAD}
\end{figure}

In \Cref{fig:fish_CAD}, we show renderings of the three types of fishtail actuators we investigate in this work. We vary material parameters as well as geometry.

\begin{figure}[tb]
    \centering
    \includegraphics[width=0.65\columnwidth]{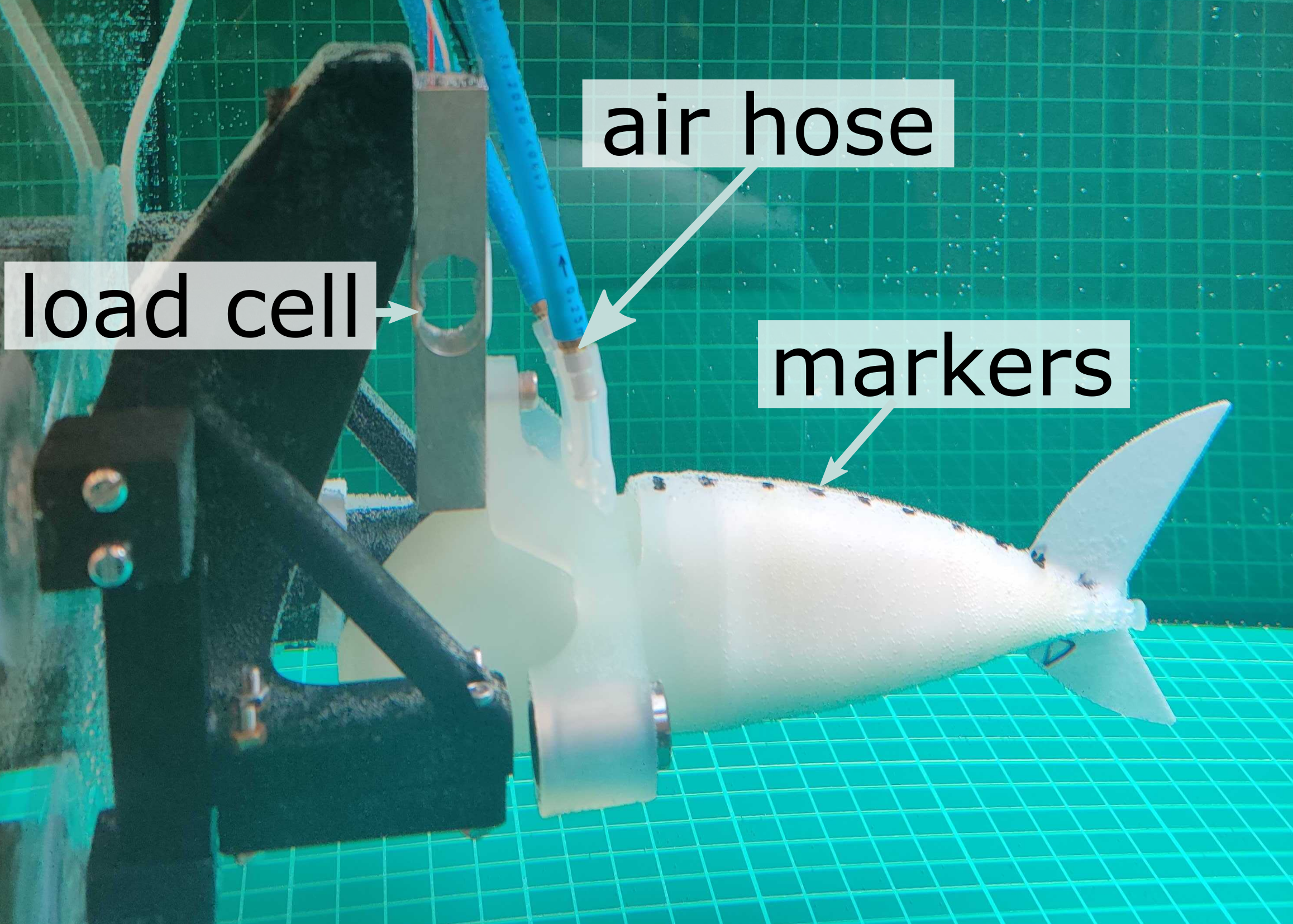}
    \caption{Thrust experiment setup actuated with a Festo proportional valve manifold. A 3D-printed fixture (black) connects the load cell (grey) to a fish adapter (opaque). The fish adapter is mounted within a casted soft robotic fish tail made of silicone elastomer. The deformation of the tail is captured using a GoPro positioned above the tail. Markers on the back of the fish are tracked using Kanade-Lucas-Tomasi (KLT) feature tracking. The thrust generated by the fish is measured using a load cell.}
    \label{fig:experimental_setup}
\end{figure}

We collect tail deformation, thrust force, and air chamber pressure data of pneumatically actuated silicone fishtail both in air and in water. To this end, the fishtail is mounted in a water tank and the fish head is rigidly connected to a load cell (TAL2210, \SI{1}{kg} model) to measure the thrust force in the heading direction (\Cref{fig:experimental_setup}). The load cell data is obtained using an amplifier board and a microcontroller.\footnote{Amplifier: HX711, Microcontroller: Arduino Leonardo} Black markers are painted onto the fish’s back and tracked using Kanade-Lucas-Tomasi (KLT) feature tracking in a video captured from above using a GoPro Hero 6 camera~\cite{lucas1981iterative,tomasi1991tracking}. We found in practice that three markers are sufficient for capturing the deformation of the fishtail and can be used as input to our learning pipeline. From the video after correcting for parallax and distortion, we obtain in-plane displacements for each marker. An LED is used to synchronize the pressure and the force data with the recorded video. The fish is actuated with a Festo proportional valve manifold with controlled pressure\footnote{MBA-FB-VI, \SI{0}-\SI{2}{bar} range, 1\% accuracy}, which outputs desired and actual pressure data for both air chamber sides of the fishtail.

\begin{table}[t]
    \centering
    \caption{Material, Young's moduli, air chambers, and maximum applied pressure for each prototype. Reported material parameters are nominal values.}
    \begin{tabular}{@{}llll@{}}
    \toprule
    Property & \emph{Nemo} & \emph{Dory} & \emph{Bruce}\\ 
    \midrule
    Body Material & DS10 & DS10 & DS20 \\ 
    Body Young's Modulus [\si\MPa] & $0.1-0.25$ & $0.1-0.25$ & $1.1$ \\
    \# of Chambers & 9 & 12 & 9 \\
    Spine Young's Modulus [\si\GPa] & $2.5-5$ & $2.5-5$ & $2.5-5$ \\
    Max. Pressure [\si\bar] & 0.2 & 0.35 & 0.5 \\
    \bottomrule
    \end{tabular}
    \label{tab:fish}
\end{table}
Two types of experiments are conducted using the same experimental setup: First, a quasistatic experiment is carried out, where the fishtail is deformed by actuating one side of the fish with constant pressure. Second, flapping experiments are conducted where the fishtail is actuated using a square pressure wave, which leads to a natural motion of the tail. Both the quasistatic and the flapping experiment are carried out with a variety of fish designs using silicone rubber with different Young's moduli.\footnote{Smooth-On Dragon Skin 10/20 slow with shore hardness 10A/20A} The fish materials, geometries, and actuation pressures are shown in \Cref{tab:fish}. No pre-straining was done on the fishtails prior to experiments and negligible hysteresis was found when actuating under periodic pressure signals. The maximum observed displacements in the fishtail were on the order of $20\%$, normalized to a tail length of \SI{10}{cm}.

Since we are not fully constraining the fish to one-dimensional translation, there exists bending moments that may result in spurious forces on the load cell. However, we assume these to be negligible compared to the peak of the thrust force. Furthermore, due to the small size of the tank, a standing wave persists in the tank even after initial actuation. This can lead to a disturbance force on the load cell. However, since the frequency of this disturbance is known, it can be filtered out during post-processing. We also note that the measurement system has its own compliance and damping, which account for the recoil and negative force readings.

%% file: main/identification.tex
\section{System Identification}
\subsection{Learning Material Parameters}
\label{system identification}
In this section, we explain our method for system identification. The process begins with exporting two STL files from a CAD file of the soft actuator, one for the body and one for the spine. To streamline the tetrahedral mesh generation, we simplify the geometry by patching the small mounting holes used during the casting process to stabilize the fish. Otherwise, the geometry of the body and spine are imported as designed without modifications. In practice, we have found that a correct chamber and spine geometry are essential for adequate system identification of material parameters that are physically plausible. 

We tetrahedralize the surface triangle mesh by modeling it as a whole piece of surface, a few holes representing the air chambers, and a set of internal points defining the separation of spine and body. The tetrahedralization is done according to the method by Hu et al.~\cite{hu2018tetrahedral}. We control the target length of edges to be $\sfrac{1}{50}$ of the whole body length to get an expressive and representative tetrahedral mesh.

After the tetrahedralization, we split the spine and body elements and assign different Young's moduli to them. The internal surfaces, which are identified as air chambers, are modeled as actuators to drive the tail. In physical experiments, we actuate the tail in the air and record its deformation through the tracked markers. This information provides supervision to the material parameter search for both the body and spine. We measure the error of matching by the Euclidean distance between measurement data and simulation results from the sensors.

We conduct our simulation experiments based on a linear elastic corotated material. The model can be improved by introducing a more sophisticated elastic energy model, for example Neo-Hookean material, which was not yet supported in our simulation framework. Furthermore, we chose not to include damping explicitly in the material model since we learn material parameters using static deformation data. For the dynamic experiments, we observed that the deformations were sufficiently small for accurate characterization and that the effect of material damping is small thus having a negligible effect on the forced response.

\subsection{Learning Hydrodynamics}
\label{learning hydrodynamics}
To learn a simplified model of the hydrodynamic thrust force predictor $f_\textrm{thrust}$, we use a feedforward neural network trained on data from the experimental setup (see appendix). The inputs of the neural network consist of the positions and velocities of the tracking points at time $t$, together with the measured pressure and its time derivative, while the output is a one-dimensional value for $f_\textrm{thrust}$ at time $t$. The feedforward neural network consists of three hidden layers with RELU activation functions and respectively a dimension of 200, 300, and 200 hidden units. The network is trained with the Adam optimizer using a learning rate of 0.001 and a mean squared error loss. Note that our hydrodynamic modeling method does not estimate fluid properties such as density or viscosity, but rather the thrust force prediction, which is of more immediate use for designers of swimming robots.

\begin{table}[h]
\caption{Specification of the training and test sets for the thrust predictor network. In the Nemo experiment, we generalize to unseen trials with seen parameters. In the Bruce experiment we generalize to unseen higher actuation frequency.}
\centering
\begin{tabular}{@{}lllll@{}}
    \toprule
    \textbf{Name} & \textbf{Mode} & \textbf{N.\ trials} & \textbf{Pressure} & \textbf{Frequency} \\
    \midrule
    \multirow{2}*{Nemo} & Training & 32 & 200,300 mbar & 1,2,3,4 Hz \\  & Validation & 8 & 200,300 mbar & 1,2,3,4 Hz \\
    \multirow{2}*{Bruce} & Training & 30 & 500,750 mbar & 1,2,3 Hz \\  & Validation & 10 & 500,750 mbar & 4 Hz \\
    \bottomrule
\end{tabular}
\label{tab:training_set}
\end{table}

In \Cref{tab:training_set}, we provide a description of the training and the test set used for the thrust predictor network. Only \emph{Nemo} and \emph{Bruce} are reported in \Cref{fig:thrust_prediction} and \Cref{tab:training_set} for brevity since \emph{Dory} thrust data predictions are similar. The table clarifies how we tested the generalizability of our predictor. We split the experimental data to show the ability of the network to generalize to unseen trials. For the fishtail named \emph{Nemo}, while we keep the parameters the same, we generalize in the validation set to unseen trials under various actuation amplitudes and frequencies. In the experiments for \emph{Bruce}, we generalize to an unseen higher actuation frequency.

%% file: main/results.tex
\section{Results}
\label{results}

\subsection{Learned Material Parameters}

\begin{figure}[t]
    \centering
    \includegraphics[width=0.85\columnwidth]{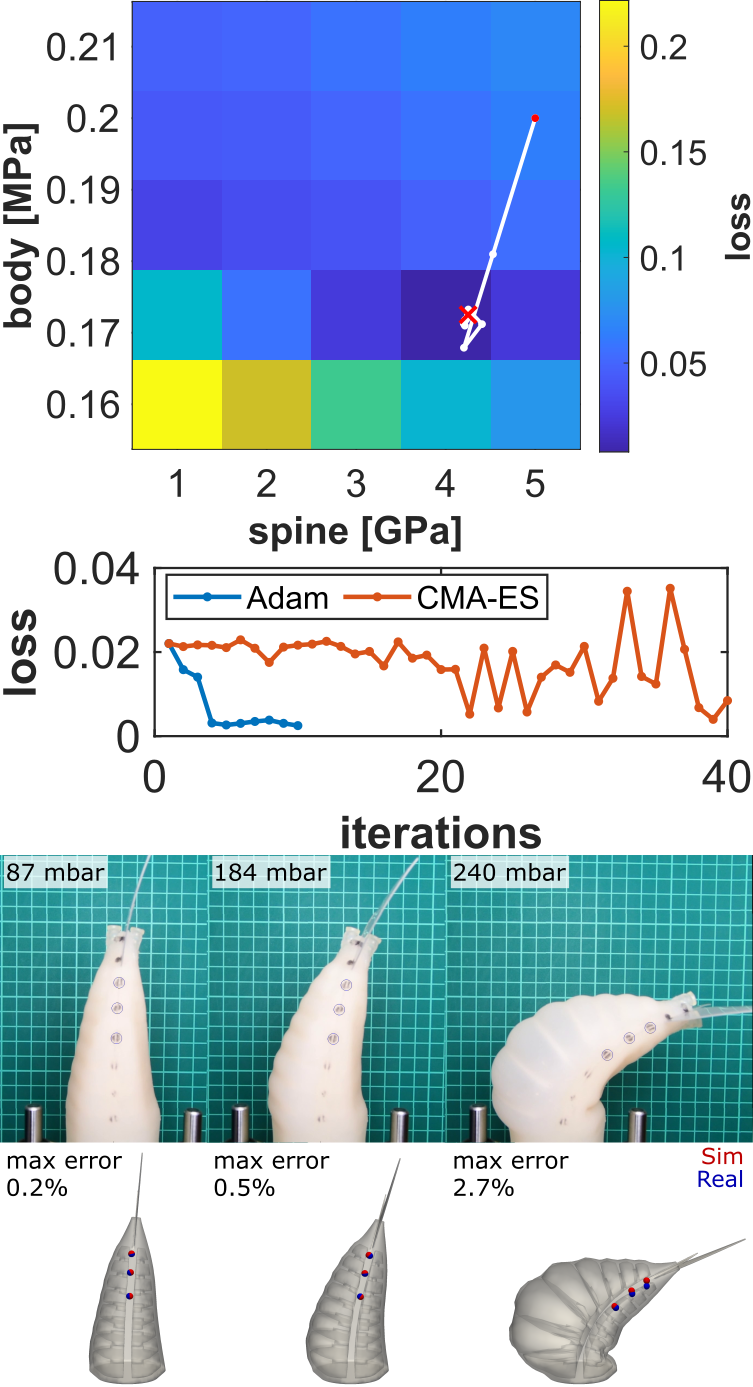}
    \caption{\textit{Top:} Exhaustive search of the Young's moduli pair for the spine and body. The moduli pair with the lowest Euclidean loss is located at the red $\times$. The red dot indicates the start of a gradient search and the white line shows the progress towards convergence. \textit{Middle:} Convergence comparison between a gradient-based (Adam) and gradient-free search (CMA-ES). \textit{Bottom:} Comparison of static deformation of the \emph{Nemo} fish for increasing pressure in experiment and simulation. Maximum displacement error increases with pressure, but remains within the measurement error (on the order of the diameter of markers). The grid in white under the real robot has \SI{10}{mm} spacing. The reported error is normalized to the fish tail length of \SI{10}{cm}.}
    \label{fig:system_id}
\end{figure}

\begin{figure}[t]
    \centering
    \includegraphics[width=0.99\columnwidth]{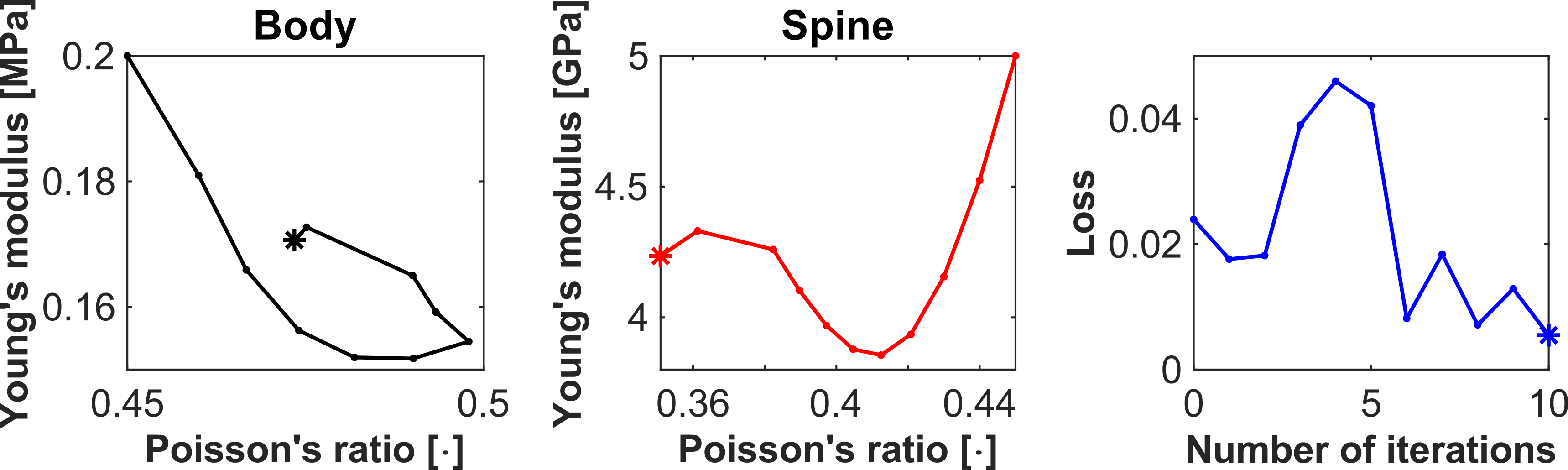}
    \caption{Learning four material parameters from deformation data take with the \textit{Nemo} fish prototype using a gradient-based approach that is run until convergence. The final values for Young's moduli and Poisson's ratios depicted as an asterisk agree with plausible material parameter values with lower loss.}
    \label{fig:4_param_search}
\end{figure}

For the \emph{Nemo} tail actuator described in \Cref{tab:fish}, we perform a grid search of the loss landscape centered around the ground truth datasheet values for the silicone and acetal materials of the body and spine. We report that if the exact geometry of the actuator is reproduced with high fidelity in the simulation, we converge to values within the range of typical measured values for the material Young's moduli (see \Cref{fig:system_id}). Further, we see that there is a unique minimum value that is within the acceptable range of measured moduli for both parameters. Note that we assume a priori that the Poisson ratio for silicone is $\nu\approx0.5$ or \textit{nearly} incompressible, a standard assumption for silicone, and the acetal sheet Poisson ratio is $\nu=0.37$ as reported by the manufacturer. Typical values for the Young's modulus of Dragon Skin 10 range from \SI{0.1}{MPa} to \SI{0.25}{MPa} and the value for Dragon Skin 20 was measured to be in the range of \SI{1.1}{MPa}~\cite{marechal2020toward} and typical values for the Young's modulus of acetal sheets range from \SI{2.5}{GPa} to \SI{5}{GPa}.\footnote{\url{https://dielectricmfg.com/knowledge-base/acetal/}}

In \Cref{fig:4_param_search}, we demonstrate that our method can be extended to higher dimensional parameter spaces such as a search over both Young's moduli and Poisson's ratios. Note that although we allow the Poisson ratio to vary in this identification experiment, the final value to which the body Poisson's ratio converges is still \textit{nearly} incompressible as expected for silicone.

\subsection{Gradient-based and Gradient-free Solver Methods}

We compare the runtime of the gradient-free method \emph{CMA-ES} against the runtime of the gradient-based method \emph{Adam} in \Cref{tab:gradient_methods}. The comparison shows that although \emph{CMA-ES}~\cite{igel2007covariance} is slightly faster in runtime per iteration, the use of the gradient-based method Adam~\cite{kingma2014adam} is significantly more effective for convergence. These comparative experiments shown in \Cref{fig:system_id} were carried out on a computer with an Intel Core i9-9900K @ 3.60GHz with 16 cores processor and 64.0 GB of memory.

\begin{table}[htb]
    \centering
    \caption{Comparison of Adam, CMA-ES, and Grid search. The forward simulation time is 318.9 seconds equivalent to grid search.}
    \begin{tabular}{@{}lllll@{}}
        \toprule
        \thead[l]{Method} & \thead[l]{Total\\Iterations} & \thead[l]{Time Per\\ Iteration} & \thead[l]{Total Time} & \thead[l]{Loss After \\4 Iterations} \\
        \midrule
        \textbf{Adam} & \textbf{4} & 334.4 s & \textbf{~22 m} & \textbf{0.0027} \\
        CMA-ES & 40 & 321.0 s & ~214 m & 0.021  \\
        Grid search & 25 & \textbf{318.9 s} & 132.9 m & 0.023 \\
        \bottomrule
    \end{tabular}
    \label{tab:gradient_methods}
\end{table}

\subsection{Dynamic Experiments}

\begin{figure}[t]
    \centering
    \includegraphics[width=0.85\columnwidth]{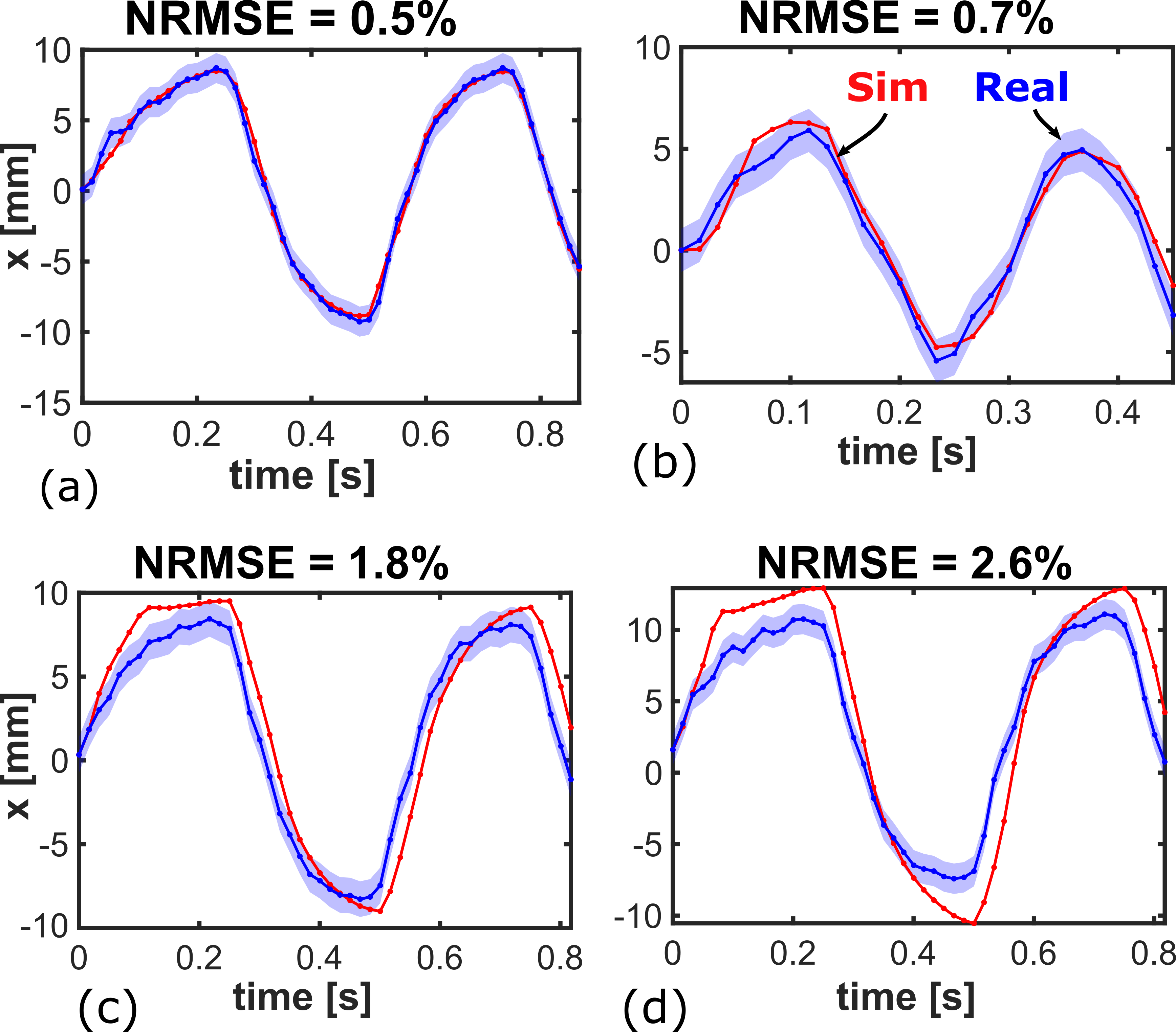}
    \caption{Simulation and measurement data for the \emph{Nemo} fish at (a) 200 mbar and 2 Hz, (b) 200 mbar and 4 Hz, (c) the \emph{Bruce} fish at 500 mbar and 2 Hz, and (d) the \emph{Dory} fish at 350 mbar and 2 Hz. The color bands indicate the variance from $N=5$ trials. For both actuation signals, we are capable of achieving sub-millimeter accuracy between experiment and simulation. The same method is used to identify the parameters of Bruce and Dory, exhibiting accuracy still within \SI{3}{mm}. We normalize the Root Mean Square Error (RMSE) to the fish tail length of \SI{10}{cm}. As expected, higher pressures result in larger deformations with greater error. The phase lag exhibited in \emph{Bruce} and \emph{Dory} may be due to actuator dynamics present during higher actuation pressures.}
    \label{fig:results}
    \vspace{-3pt}
\end{figure}

For dynamic experiments, we compare the results of our simulation output with the measured data of the furthest tracked dot on the tail. In Fig.~\ref{fig:results}, we report our findings for sim2real performance in both the standard fish (\emph{Nemo}) and two other data sets for a tail with different Young's moduli for the body (\emph{Bruce}) and a tail with a greater number of air chambers (\emph{Dory}). We demonstrate that if system identification is done correctly our simulation results can predict the performance of a novel actuator design to within millimeter precision or $3\%$ max normalized error using only a quasistatic data set for training without need for material testing. The \textit{Bruce} prototype required higher pressures to get similar displacements to \textit{Nemo} since the material is stiffer. The prototype \textit{Dory} required higher pressure as well to produce similar displacements due to a greater number of actuation chambers. For videos of the dynamic experiments and simulation, we ask readers to refer to our supplemental video.

\subsection{Learned Hydrodynamics}
\label{sec:learned_hydrodynamics}
\begin{figure}[t]
    \centering
    \includegraphics[width=1\columnwidth]{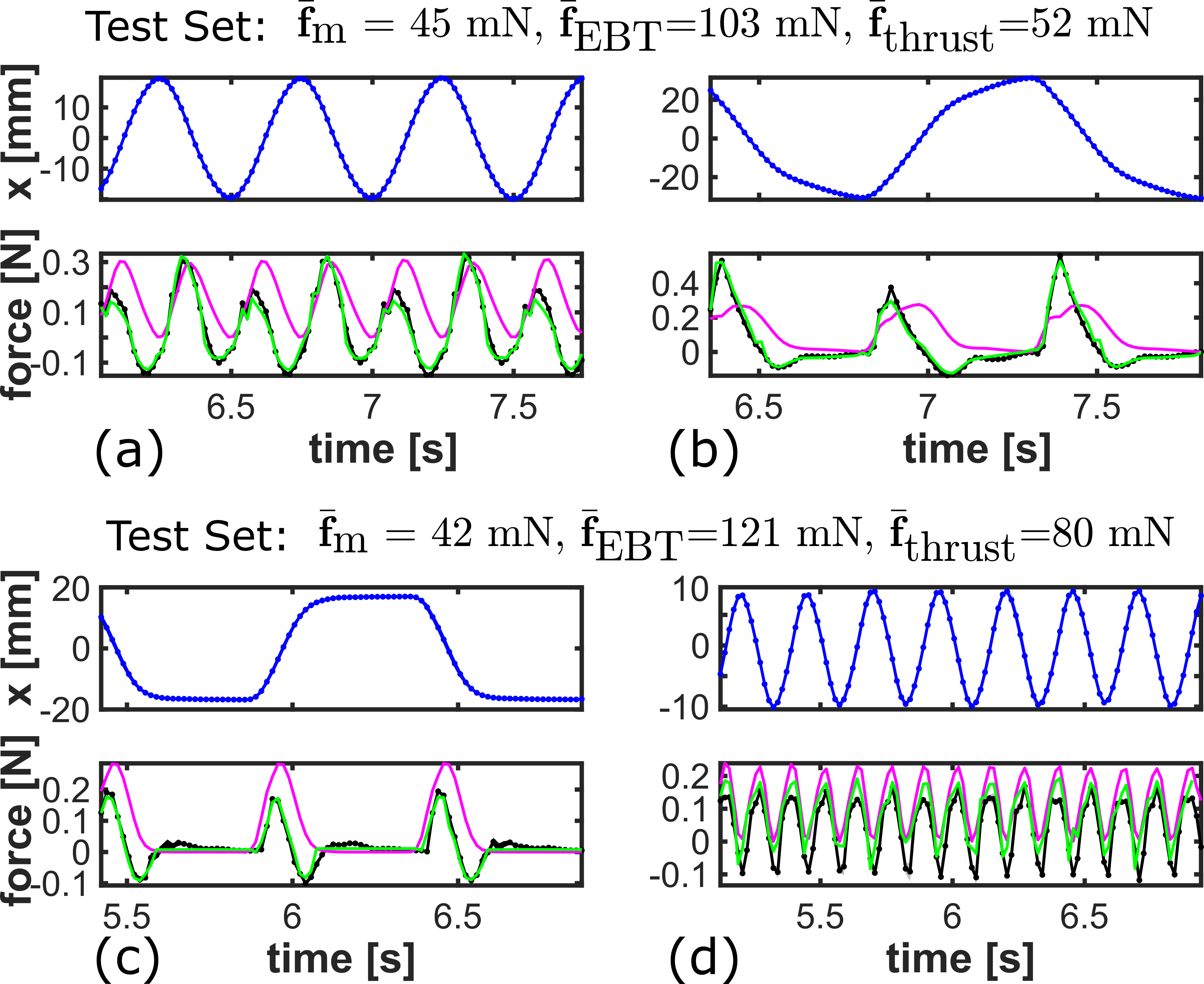}
    \caption{Tail lateral displacement (blue) with thrust measurement and prediction. We compare the measured thrust force $\mathbf{f}_\textrm{m}$ (black), analytical thrust force from EBT $\mathbf{f}_\textrm{EBT}$ (magenta), and the neural network thrust prediction $\mathbf{f}_\textrm{thrust}$ (green). We show the thrust prediction for two training sets (a) and (c) and we report the time-averaged thrust for the two test cases (b) and (d). We note that EBT tends to over-predict the thrust measurement while our neural network thrust prediction accurately reproduces the frequency for a given actuation signal and matches amplitude more robustly than EBT.}
    \label{fig:thrust_prediction}
\end{figure}

We compare our simple predictor of thrust force with the measurement data and the theoretical thrust from EBT, a classic, non-learning-based approximate analytical model from the literature, in~\Cref{fig:thrust_prediction}. Although the model is capable of generalizing to actuation signals at frequencies previously unobserved in the training for the \emph{Nemo} prototype, for the \emph{Bruce} prototype the discrepancy is large, nearly twice the measured force, likely because of the more limited training data.
In comparing to the thrust prediction from EBT, we see that the analytical thrust is strictly positive. The negative measured force is due to recoil of the measurement system. The advantage of our approach is the differentiability and flexibility of the neural network provided enough data is collected. We note that the network is capable of learning the measurement dynamics due to the compliance of the load cell and the damping of the water though these effects can be considered separately and corrected for if desired (see \Cref{sec:load_cell_measurement}). We conjecture that the asymmetry of the measured thrust may be due to imperfections in the fabrication process of the fish tails favoring one direction more strongly.

%% file: main/conclusion.tex
\section{Conclusion}
\label{conclusion}
We present an experimentally-verified simulation framework that can be used to accurately predict the deformations of a pneumatically actuated fish tail with a flexible spine.
Our pipeline can accurately learn material parameters from a quasi-static data sets without having to do expensive and time-consuming material testing. It also eliminates the need to do manual tuning of material constants to get accurate simulation results. The parameters we found are not only within typical range of measured material parameters for our materials, but can be used to successfully predict the behavior of dynamic experiments for different pressure actuation amplitudes and frequencies to within $3\%$ positional error normalized to a actuator length of \SI{10}{cm}. Although we use an isotropic corotated material, which is linear elastic, we find that this model is more sufficient to model large deformations on average giving acceptable displacement results for our engineering application. In these experiments, the damping of the material and the hydrodynamic effects are found to be negligible. This is because the actuation pressures used dominate the deformation compared to losses and hydrodynamic pressure. 

We show a data-driven approach can be used to do simple prediction on a useful performance metric such as thrust force given a suitable hardware setup. However, more work is needed to produce a more robust thrust predictor if the morphology of the actuator changes substantially. We claim that for small design changes such as the choice of silicone or the number of internal chambers this framework can be used to quickly assess the relative merits of each design with a relatively sparse data set of approximately 30 types of different actuation signals.

Our aim is to further progress towards a systematic method by which soft roboticists can simulate and optimize their designs and controllers, whether they be soft fish, manipulators, or other flavors of soft robots. A fast and physically-verified co-optimization method of design and control is the goal.

%% file: main/appendix.tex
\section{Appendix}

\subsection{Large Amplitude Elongated-Body Theory}
\label{sec:large_amplitude_ebt}
The thrust of a fish swimming in an inviscid fluid can be described by Lighthill's Elongated-body theory (EBT)~\cite{lighthill1971large}. To calculate the thrust predicted by EBT, we make use of a simplified version of the reaction force expression from Lighthill~\cite{lighthill1971large}, which uses significantly different notation,
\begin{equation}
    f_\textrm{EBT} \approx \frac{1}{2}m\dot{x}^2,
\end{equation}
where we have evaluated the expression for the marker closest to the tail and we have assumed the unit vectors $\frac{\partial x}{\partial a}\approx 0$ and $\frac{\partial y}{\partial a} \approx 1$ when considering small tail deformations in dynamic actuation. Note that $\dot{x}$ is velocity tangential to the fin chord.

We also use Lighthill's expression for the virtual mass $m=\frac{1}{4} \pi \rho s^2$, where $\rho$ is the density of the fluid and $s$ is the cross-sectional depth, by estimating the geometry from the CAD model of our robotic fish.

\subsection{Load Cell Measurement Dynamics}
\label{sec:load_cell_measurement}

\begin{figure}[htb]
    \centering
    \begin{circuitikz}
    
    \pattern[pattern=north east lines] (0,0) rectangle (.25,3);
    \draw[thick] (.25,0) -- (.25,3);
    
    \draw (.25,1) to[spring, l_=$k$] (2,1);
    \draw (.25,2) to[damper, l=$b$] (2,2);
    \draw[fill=gray!40] (2,0.5) rectangle (4,2.5);
    \node at (3,1.5) {$m$};
    
    \draw[thick, ->] (5,1.5) -- (4,1.5);
    \node at (4.5,2) {$f_\textrm{hydro}$};
    
    \draw[thick, ->] (0,1.5) -- (-1,1.5);
    \node at (-0.5,2) {$f_\textrm{m}$};
    
    \end{circuitikz}
    \caption{Load cell dynamics. The force $f_\textrm{m}$ is measured with a pre-calibrated load cell. The force $f_\textrm{hydro}$ is the actual hydrodynamic force experienced by the fish. The stiffness $k$ and the damping coefficient $b$ describe the lumped mechanical impedance of the measurement system. The lumped mass $m$ is dominated by the fish. The wall is the connection point to the load cell.}
    \label{fig:load_cell_dynamics}
\end{figure}
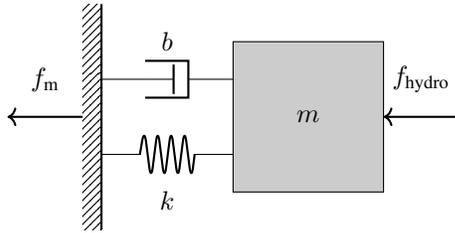

We choose to model the dynamics of the load cell measurement jig (\Cref{fig:experimental_setup}) as a lumped element system shown in \Cref{fig:load_cell_dynamics}. We assume that the stiffness of the fixture is far greater than the stiffness of the connection between the load cell and the fish. 
Solving for the transfer function describing the load cell dynamics, we have
\begin{equation}
    \frac{f_\textrm{m}}{f_\textrm{hydro}} = \frac{bs+k}{ms^2+bs+k},
    \label{eqn:measurement_dynamics}
\end{equation}
where $s$ is the complex Laplace variable. From this model, it is clear how the measured force $f_\textrm{m}$ can be negative even for strictly positive $f_\textrm{hydro}$ as discussed in \Cref{sec:learned_hydrodynamics}. 

It is possible to measure each parameter in \Cref{eqn:measurement_dynamics} and compensate it by inverting the transfer function, however, our proposed data-driven neural network force predictor is agnostic to the dynamics of the measurement system.

